\documentclass[10pt, twocolumn, letterpaper]{article}

\usepackage{cvpr}
\usepackage{times}
\usepackage{epsfig}
\usepackage{graphicx}
\usepackage{amsmath}
\usepackage{amssymb}
\usepackage{textcomp}
\usepackage{gensymb}
\usepackage{multirow}
\usepackage{float}
\usepackage{xcolor}

\usepackage[pagebackref=true,breaklinks=true,colorlinks,bookmarks=false]{hyperref}

\cvprfinalcopy 

\def\cvprPaperID{6496} 

\ifcvprfinal\fi
\begin{document}

\title{GroundNet: Monocular Ground Plane Normal Estimation \\ with Geometric Consistency}

\author{Yunze Man$^{1}$\thanks{Work done when Yunze was an intern at CMU.} , Xinshuo Weng$^{2}$, Xi Li$^{1}$, Kris Kitani$^{2}$
\\
Dept. of Computer Science and Technology, Zhejiang University, Zhejiang, China$^{1}$ \protect\\ Robotics Institute, Carnegie Mellon University, Pittsburgh, PA, USA$^{2}$ \protect\\ \{yzman,xilizju\}@zju.edu.cn, \{xinshuow, kkitani\}@cs.cmu.edu
}

\maketitle

\begin{abstract}
\vspace{-0.3cm}
We focus on estimating the 3D orientation of the ground plane from a single image. We formulate the problem as an inter-mingled multi-task prediction problem by jointly optimizing for pixel-wise surface normal direction, ground plane segmentation, and depth estimates. Specifically, our proposed model, GroundNet, first estimates the depth and surface normal in two separate streams, from which two ground plane normals are then computed deterministically. To leverage the geometric correlation between depth and normal, we propose to add a consistency loss on top of the computed ground plane normals. In addition, a ground segmentation stream is used to isolate the ground regions so that we can selectively back-propagate parameter updates through only the ground regions in the image. Our method achieves the top-ranked performance on ground plane normal estimation and horizon line detection on the real-world outdoor datasets of ApolloScape and KITTI, improving the performance of previous art by up to 17.7\% relatively.
\end{abstract}

%
\maketitle

\vspace{-0.4cm}

\section{Introduction}
\vspace{-1mm}
Estimating the 3D orientation of the ground plane is an important pre-processing step for ground robots \cite{Manglik2019}, wearable camera systems \cite{Weng2018_r2n} and autonomous driving \cite{Wang2018}. An accurate estimate of the ground plane can serve as an important prior for many perception and planning tasks, \eg, 3D object detection \cite{Chen2015, Ku2018, Weng2019}, 3D object tracking \cite{Khan2006, Weng2019_3dmot}, 3D semantic segmentation \cite{Alvarez2012}, image data synthesis \cite{Lee2016}, free space estimation \cite{Pfeiffer2010}, camera placement estimation \cite{Zhang2003}, 3D reconstruction \cite{Hoiem2007}, and scene analysis \cite{Hoiem2006, Hedau2010}. While many sensors can be used to directly estimate the ground plane (\eg, depth camera, stereo camera, laser scanner), we are primarily interested in innovating ground plane normal estimation for mobile platforms that are equipped with only a single RGB camera.

Perhaps the most classical approach to ground plane estimation makes use of multi-view geometry or motion cues to first triangulate points in 3D or directly obtain the 3D point cloud using depth sensors like LIDAR. Then a large plane is fitted to the 3D points using a robust model fitting algorithm like RANSAC \cite{McDaniel2010}. When only a single image is available, parallel lines detected on the ground plane can be used to estimate vanishing points and the horizon line \cite{Hartley2007}. While these geometry-based approaches are exact with noiseless input, their performance is highly dependent on the reliability of low-level computer vision algorithms to extract corner points or line segments. Not to mention that outdoor scenes typically lacks 3D architectural prior (\ie, the vertical relationships between walls, floor, and ceiling), making existing geometry based method insufficient for the estimation.

An alternative approach that can be used to estimate the ground plane is the use of monocular surface normal estimation \cite{Bansal2017}. The basic idea of this approach is to use small image patches to estimate the distribution of surface normals using local visual information. Most prior work in this area are pixel-wise normal estimation models. They are tailored to indoor scenes and are typically not designed to deal with the outdoor scenes which involve heavy noise, including diverse objects (\eg, vehicles, pedestrians) and natural `stuff' (\eg, vegetation, sky) with dynamic motion. For this reason, the estimated normals often suffer from heavy local noise.

To leverage the advantages of both approaches, we propose the GroundNet, which computes the ground plane normal using the depth and surface normal cues in two separate streams: (1) in the surface normal estimation stream, we obtain a pixel-wise surface normal map of the ground region and then take the average over pixels as the ground plane normal; (2) in the depth estimation stream, we lift the ground regions in the input image to a point cloud using the estimated depth. Then a ground plane is fitted to the point cloud using the differentiable RANSAC (DSAC) \cite{Brachmann2017}.

To remove the effect caused by diverse objects nearby the ground in outdoor scenes, a ground segmentation stream is used to isolate the ground regions. We then selectively back-propagate parameter updates only through the ground regions in the image. We share the backbone encoder network for all three networks: ground region segmentation, depth estimation and pixel-wise normal estimation.

In addition, our insight is that the estimated depth and surface normal should be geometrically consistent, meaning that the computed ground plane normals from the depth and surface normal streams should be nearly the same. However, this is not necessarily true in reality. To introduce this geometric consistency between the depth and surface normal streams, we propose to add a consistency loss to minimizes the angular difference between the computed ground plane normals in two streams. This supervision signal is flown back to the surface normal and depth estimation streams such that our GroundNet can implicitly learn to predict the geometrically-consistent depth and surface normal in two streams. We argue that this consistency constraint can resolve the geometrical ambiguity when estimating the depth or normal alone from a single image.

\begin{figure*}[t]
  \centering
  \setlength{\abovecaptionskip}{-0.5cm}
  \setlength{\belowcaptionskip}{-0.3cm}
  \includegraphics[trim=1cm 1cm 1cm 1.9cm, clip=true, width=0.9\textwidth]{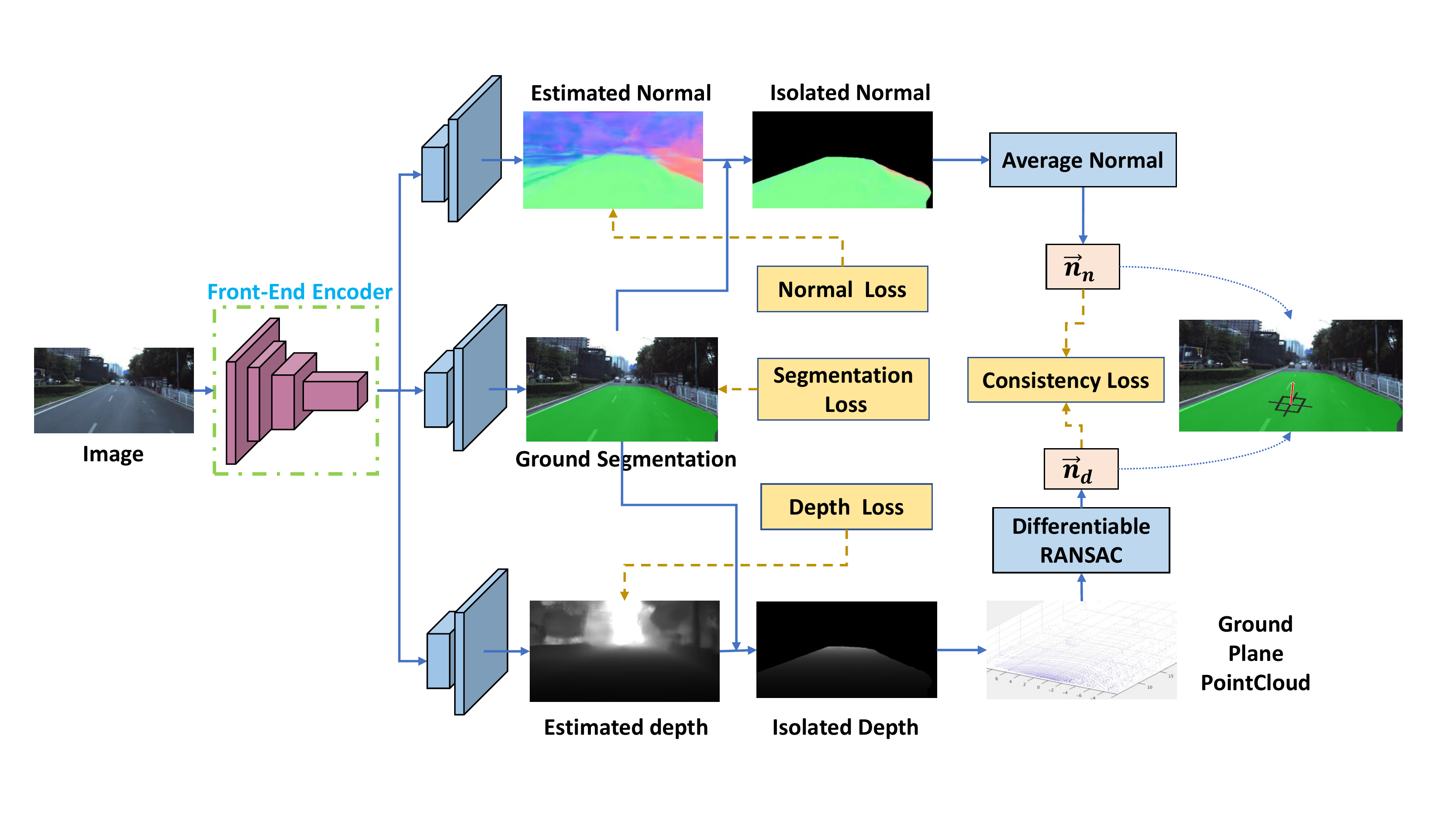}
  \vspace{-0.5cm}
  \caption{GroundNet: a single image is fed into the front-end encoder and then outputs three streams, namely surface normal estimation, depth estimation and ground segmentation. (1) The ground segmentation is used to isolate the ground regions so that we can selectively back-propagate parameter updates through only the ground regions in the image. (2) Depth stream: we convert the ground region in the image into a point cloud using the estimated depth, and then compute the ground plane normal $\mathbf{n}_d$ by fitting a plane to the point cloud using the Differentiable RANSAC module. (3) Normal stream: we calculate the ground plane normal $\mathbf{n}_n$ by taking the average of surface normal vectors of all pixels in the ground region. The final output of our ground plane normal is the average of the ground plane normals computed from the depth and normal streams.
  }
  \vspace{0.25cm}
  \label{main}
\end{figure*}

The proposed GroundNet is shown in Figure \ref{main}. To show the effectiveness of our method, we evaluate on the real-world outdoor datasets of ApolloScape and KITTI, which involves heavy noise from objects. As the images in these two datasets lack of variety of the ground plane normal, we augment the images by adding random roll and pitch so that estimating the ground plane normal becomes challenging. Our method achieves the state-of-the-art performance on the datasets either with or without the augmentation.

Our contributions are summarized as follows: (1) we propose a novel network for end-to-end outdoor ground plane normal estimation; (2) we introduce a ground segmentation stream to isolate the ground regions so as to avoid the noise caused by irrelevant objects in outside scenarios; (3) we propose to learn the essential correlation between depth and normal by geometric consistency loss, allowing better learning of 3D information through multi-modality interaction and refinement. (4) though extensive evaluation on the real-world outdoor datasets of KITTI and ApolloScape, the proposed GroundNet achieves the state-of-the-art performance on the task of ground plane normal estimation and horizon line detection, improving the previous state-of-the-art by up to 17.7\% relatively.
\vspace{-1mm}

\section{Related Work}
\vspace{-1mm}
\noindent\textbf{Ground Plane Normal Estimation.} Existing methods for ground plane estimation can be classified into geometry-based methods and learning-based methods. Geometry-based methods often extract the 3D scene structure (\eg, using multi-view cues, motion cues or depth sensors) and then the ground plane is fitted to the 3D points using a robust model fitting algorithm like RANSAC. \cite{McDaniel2010} identifying the ground using the 3D point cloud from LIDAR. \cite{Mufti2012} obtains the video frame rate depth maps from the time-of-flight (TOF) cameras and exploits 4D spatiotemporal RANSAC for ground plane estimation. \cite{Se2002} generates the 3D point cloud under a stereo setup and then estimates the ground plane by the disparity. Assuming the scene is static, simultaneous localizing and mapping (SLAM) and structure from motion (SFM) approaches can also be used for extracting the 3D scene structure \cite{Tardif2008, Micusik2009, Schonberger2016, Ovren2018, Zhou2012, Yuan2006}, making ground plane estimation possible. 

When only a single image is available, parallel lines detected on the ground plane can be used to estimate vanishing points and the horizon line \cite{Hartley2007}. Also, \cite{Micusik2008} shows how grouping detected line segments into quadrilaterals can be used to find orthogonal planes. However, these geometry-based methods are highly dependent on the reliability of low-level computer vision algorithms (\eg, planar homography estimation, line segment detection), especially when given only a single image.

The secondary category methods focus on applying machine learning technique to estimate the ground plane normal either directly or indirectly from related tasks. There are only a few prior works are direct methods. \cite{Haines2012, Haines2015} learn a classifier to classify local planar image patches and their orientations first. Then a Markov random field (MRF) \cite{Weng2018_imagelabeling} model is learned to segment the image into dominant plane segments. \cite{OsunaCoutino2016} achieves the ground plane recognition by learning the lighting-invariant texture feature using regularized logistic regression model. However, these methods make use of shallow learning model and do not benefit from the recent significant progress in deep learning. Also, they do not model the geometric relationship existing in the image. To the best of our knowledge, GroundNet is the first work of direct ground plane estimation which leverages the strong capacity of the deep neural network and the geometry consistency. 

Also, one can estimate the ground plane implicitly by solving a related task. One such example of the task is 3D surface layout recovery \cite{Hoiem2007, Zhang2014, Zou2018}. These methods are capable of creating a simple indoor layout reconstruction from a single image and then the ground plane normal is able to be estimated. Another example task could be monocular surface normal estimation \cite{Ren2018, Chen2017, Bansal2017, Bansal2016, eigen2015predicting, Li2015, Wang2015, Wang2016}, which usually formulate the problem as a dense pixel-wise prediction problem and learn a feed-forward deep neural network classification model. On top of the estimated pixel-wise normals, one can group the pixels with similar normals into the dominant planes and then compute the average normal for each plane. Although these methods achieve significant progress, most of them are tailed only for the indoor scene while our method is not limited to the scene and can work significantly well on the real-world outdoor datasets. In addition, since these prior works do not explicitly estimate the plane normal, the fact that the ground plane is often a flat and smooth surface is ignored. In contrast, we parameterize the output of our methods to be a planar surface normal explicitly and in the meantime leverage the successful architecture design from the surface normal estimation methods.

\vspace{0.2cm}
\noindent\textbf{Self-Supervised Learning via Geometric Consistency.} Geometric consistency is proved to be a useful and free supervision signal in many tasks. \cite{Zhou2016} learns to predict dense flow field between different instances of the same category object consistently across the synthetic and real domain. \cite{Weng2018, Wang2019, Wang20192} propose the cycle consistency loss as a free supervision to learn a tracker. \cite{Qian2016} jointly optimizes the 3D surface point positions and normals to be consistent with the observed light refraction effect. \cite{Dong2018} learn the keypoint detector, which is consistent across either different viewpoints or adjacent frames. In order to force the predicted 3D bounding box to be consistent with its 2D proposal, \cite{Weng2019} proposes a 2D-3D bounding box consistency loss.
\cite{Brostow2017} learns a depth estimator under a stereo setting by enforcing estimated depth in two views to be consistent with the disparity. Also, the consistent camera pose estimator and 3D shape predictor are learned via the multi-view projection consistency loss in \cite{Tulsiani2018}. \cite{Cheng2018} learns a 3D geometrically-consistent feature map for reconstruction, segmentation and classification from multi-view observations. \cite{zhou2017unsupervised, Mahjourian2018} enforces the estimated ego-motion consistent with the computed motion using 3D (iterative closest point) ICP on the estimated depth between two frames.

Perhaps \cite{Yang2018, qi2018geonet, Yang2018_AAAI} are closest to our work in the aspect of predicting geometrically consistent depth and surface normal. It has been shown in prior works that the depth and normal are complementary and thus jointly optimizing the two with consistency constraints can improve the performance on both tasks. In our proposed method, we also propose to leverage this consistency loss. Different from prior works, the consistency loss is applied to the estimated ground plane normals from two streams (\ie, the normal estimation stream and depth estimation stream).

\vspace{-0.2cm}
\section{Geometric Definition} \label{parameterization}
\vspace{-0.1cm}
\noindent\textbf{Parameterization.} Given a single RGB image as input, our goal is to estimate the 3D orientation of the ground plane if existing in the image, which is usually represented by an normal vector $\mathbf{n}=[n_x,n_y,n_z]$ in the world coordinate. In addition, we can also represent the normal vector $\mathbf{n}$ as $(\theta, \psi)$, where $\theta$ and $\psi$ are the roll and pitch with respect to the up-axis. Furthermore, we can convert the $(\theta, \psi)$ representation into a horizon line representation $(\theta, \rho)$, where $\theta$ is the angle that the horizon line makes with the horizontal axis, and $\rho$ is the perpendicular distance from the principal point to the horizon line. This allows us to compare our method against other horizon line estimation methods (see Sec. \ref{Metrics}).

\vspace{2mm}
\noindent\textbf{Pinhole Camera Model.} In this paper, we assume the images are captured by a pinhole camera. Suppose $[u_i, v_i]^T$ is the location of a 2D point $P_i$ on the image plane and $Q_i=[x_i, y_i, z_i]^T$ is the corresponding 3D location in the camera coordinate, we have
\begin{equation} \label{eq:project}
\begin{aligned}
\vspace{-0.3cm}
& x_i=(u_i-c_x) * z_i/f_x, \\
& y_i=(v_i-c_y) * z_i/f_y,
\end{aligned}
\end{equation}
where $[c_x, c_y]^T$ represents the principle point on the image plane and $z_i$ is the depth of the point. $f_x$ and $f_y$ are the focal length along the $x$ and $y$ directions respectively. It can be also written in homogeneous coordinate as
\vspace{-0.1cm}
\begin{align} \label{eq:pinhole}
    P_i & = \begin{bmatrix}
            \lambda u_i \\ 
            \lambda y_i \\
            \lambda
            \end{bmatrix}
            =
            \begin{bmatrix}
            f_x & 0 & c_x\\
            0 & f_y & c_y\\
            0 & 0 & 1
           \end{bmatrix}
           \begin{bmatrix} x_i \\ y_i \\ z_i\end{bmatrix}
           = K_cQ_i
\end{align}
where $K_c$ is intrinsic matrix. From Eq. \ref{eq:project} and \ref{eq:pinhole}, we know that given a 2D point $P_i=[u_i,v_i]$, the depth in camera coordinate $z_i$ and camera intrinsic matrix $K_c$, we can calculate its 3D location in the camera coordinate $Q_i=[x_i,y_i,z_i]$. We will use this formulation later in depth stream to convert the image into a point cloud.


\vspace{-0.2cm}
\section{Method} \label{Method}
\vspace{-0.1cm}
\subsection{Approach Overview} \label{Overview}

Figure \ref{main} depicts the network architecture of the proposed GroundNet. Given a single input image, a front-end fully convolutional encoder first outputs three streams, including a surface normal estimation stream, depth estimation stream and a ground segmentation stream. (See Sec. \ref{Implementation} for detailed encoder and stream design.) The segmented ground is used to isolate the ground region in the estimated depth and surface normal, getting rid of the irrelevant objects and noises. We then compute the ground plane normals from the isolated surface normal and depth estimates. 

So far, the problem has been formulated as a multi-task prediction problem, including depth estimation, normal estimation and an auxiliary ground segmentation task. However, a vital information is left unused --- the geometric correlation between depth and normal. While multi-task learning provides us with informative features and enables cross-modality interactions, geometric consistency is the core constraint that allows the multi-modality information to mutually and correctly refine the prediction. To enforce explicit geometric consistency during training, we add a consistency loss between two ground plane normals computed from the depth and normal streams. During inference, the final output of our ground plane normal is the average of the ground plane normals computed from two streams, although they are already very close after training with the geometric consistency loss.

\vspace{-0.1cm}
\subsection{Depth Estimation Stream} \label{DStream}
\vspace{-0.1cm}
Given an image $I$, we first estimate the pixel-wise depth $D(I)$ on the entire image. Then, a ground mask $M(I)$ is used to remove the non-ground region from the estimated depth $D(I)$,
\begin{equation}
    \hat{D}(I)=D(I) \odot M(I)
\end{equation}
As a result, $\hat{D}(I)$ contains only depth of the ground region, which we call the isolated depth. Following Eq. \ref{eq:project} and Eq. \ref{eq:pinhole}, we can project every pixel $i$ in $\Hat{D}(I)$ to 3D space as $Q(x_i, y_i, z_i)$ given the depth $\hat{D}(i)$ to obtain a 3D point cloud $C$ for points on the ground. We then use a plane fitting algorithm $f(\cdot)$ to get the ground plane normal $\mathbf{n}_d=f(C)$ from the point cloud, where function $f(\cdot)$ can be either least square module, as used in \cite{qi2018geonet}, or RANSAC-based module. 

\vspace{0.2cm}
\noindent \textbf{Least Square (LS) Module.} When using the LS module, it is typically assuming that all points in the point cloud $C$ lie on the same plane without any outlier point. Therefore, 
\begin{equation}
    \mathbf{Cn}=\mathbf{b}
\end{equation}
where $\mathbf{n}$ is the normal vector of the plane, $\mathbf{b} \in \mathcal{R}^{n\times3}$ is a constant vector. In the least square equation, we want to minimize $\left \| \mathbf{Cn}-\mathbf{b}\right \|^2 $, which leads to a close-form solution as below

\vspace{-0.2cm}
\begin{equation} \label{eq:square}
    \mathbf{n}=\frac{(\mathbf{C}^T\mathbf{C})^{-1}\mathbf{C}^T\mathbf{1}}{\left \| (\mathbf{C}^T\mathbf{C})^{-1}\mathbf{C}^T\mathbf{1} \right \|}
\end{equation}
where $\mathbf{1} \in \mathcal{R}^K$ is a constant vector with all 1 elements. Eq. \ref{eq:square} is differentiable and thus can be used in neural network for gradient propagation.

\vspace{0.2cm}
\noindent \textbf{RANSAC-based module.} To get rid of the adverse effect from the outliers, it is more suitable to use RANSAC than the LS. As the original RANSAC method is non-differentiable and cannot be used in neural network, we use the differentiable RANSAC (DSAC) proposed in \cite{Brachmann2017}. Instead of using deterministic hypothesis selection ($argmax$), DSAC proposes to use a probabilistic selection, \emph{i.e.}, the probability of selecting a hypothesis is higher when the hypothesis contains more inliers and has a higher score. We show in Table \ref{eval_normal} in the experiment section that differentiable RANSAC module out-performs the lease square module as the differentiable RANSAC module is robust to outliers.

\subsection{Surface Normal Estimation Stream} \label{NStream}
In general, surface normal estimation methods suffer from diverse objects when applied to outdoor scenes. Therefore, similar to the depth stream, a ground mask $M(I)$ is used to isolate the ground region from the entire normal map $N(I)$, which we estimate from the input image $I$. 

\vspace{-0.3cm}
\begin{equation}
    \hat{N}(I)=N(I) \odot M(I)
\end{equation}
As a result, $\hat{N}(I)$ contains the normal only for pixels on the ground, which we call the isolated normal map. Then, to compute the ground plane normal (\emph{i.e.}, a single normal vector representing the ground plane) from the pixel-wise isolated normal map, we simply take the average of normal vectors of all pixels on the ground in $\hat{N}(I)$:

\vspace{-0.3cm}
\begin{equation}
    \mathbf{n}_n=\frac{1}{M}\sum_{p_i \in M(I)}{\Hat{N}(p_i)}
    \vspace{-0.1cm}
\end{equation}
where M is the total number of pixels on the ground in $\hat{N}(I)$ and $\mathbf{n}_n$ is the final output of the ground plane normal computed from the normal estimation stream.

\begin{figure}[!tb]
  \centering
  \setlength{\abovecaptionskip}{0cm}
  \setlength{\belowcaptionskip}{0cm}
  \includegraphics[width=0.48\textwidth]{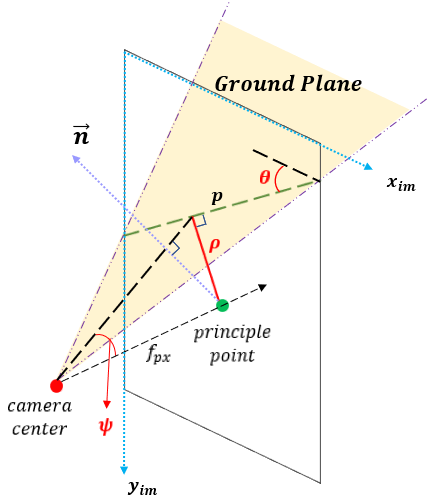}\\
  \vspace{-0.1cm}
  \caption{Illustration of the camera model. Ground plane normal $\vec{\mathbf{n}}$ can be transformed into image horizon line, parameterizing as roll $\mathbf{\theta}$ and pitch $\mathbf{\rho}$, where $\rho = f_{px} \cdot tan\psi$. 
  }
  \label{math}
\end{figure}

\subsection{Geometric Consistency} \label{Consistency}
From the geometric perspective, depth and surface normal are strongly correlated, which follows local linear orthogonality. Formally, for each pixel $p_i$, such a correlation can be written as a quadratic minimization for a set of linear equations:

\vspace{-0.5cm}
\begin{equation} \label{eq:consistency}
\mathbb{E}\left(\hat{D}(p_i), \hat{N}(p_i)\right)=\sum_{p_{j} \in \Omega(p_i)}\left\|(Q_i-Q_j)^{T} \hat{N}(p_i)\right\|^{2}
\vspace{-0.3cm}
\end{equation}
where $\Omega(p_i)$ is the set of neighbors of $p_i$, and $Q_i$, $Q_j$ are 3D points projected from 2D points $p_i$, $p_j$ given their depth in $\hat{D}$. In addition, in this work, we assume that the ground is a single flat plane. Therefore, Eq. \ref{eq:consistency} will lead to the same result for any pixel $p_j$ on the ground. In other words, the equation holds even when the neighborhood is the entire ground region, \ie $\Omega(p_i) = M(I)$. In this way, $\hat{N}(p)$ turns to $\mathbf{n}_n$. Also, $\mathbf{n}_d$ represents the plane fitted over all 3D points (Eq. \ref{eq:square}). Therefore, minimizing Eq. \ref{eq:consistency} is equivalent to minimize the difference between $\mathbf{n}_d$ and $\mathbf{n}_n$.

\subsection{Loss Functions} \label{Loss}
We now explain the loss functions of GroundNet. For pixel $p_i$, we denote the isolated depth map as $\Hat{D}(p_i)$ and ground truth depth for the ground region as $D^{gt}(p_i)$ respectively. Similarly, we denote the isolated surface normal map as $\Hat{N}(p_i)$ and ground truth surface normal for the ground region as $N^{gt}(p_i)$. Then, for the depth estimation, the loss $\mathcal{L}_{depth}$ is expressed as
\vspace{-0.1cm}
\begin{equation}
    \mathcal{L}_{depth}=\frac{1}{M}\sum_{p_i \in M(I)}{\left \| \Hat{D}(p_i) - D^{gt}(p_i) \right \|_2^2}
    \vspace{-0.1cm}
\end{equation}
where $M$ is the total number of pixels in isolated depth map. Similarly, the surface normal estimation loss $\mathcal{L}_{normal}$ is:

\vspace{-0.3cm}
\begin{equation}
    \mathcal{L}_{normal}=\frac{1}{M}\sum_{p_i \in M(I)}{\left \| \Hat{N}(p_i) - N^{gt}(p_i) \right \|_2^2}
\end{equation}

In addition, instead of predicting the depth and surface normal map independently, we enforce the geometric consistency between them by adding a consistency loss $\mathcal{L}_{con}$ on top of the ground plane normals computed from depth and surface normal streams: 

\vspace{-0.2cm}
\begin{equation}
    \mathcal{L}_{con}=\arccos{\frac{\mathbf{n}_d \cdot \mathbf{n}_n}{\lVert \mathbf{n}_d \rVert \cdot \lVert \mathbf{n}_n \rVert}}
\end{equation}

Then, the overall loss function for the GroundNet is the weighted sum of all loss terms,
\vspace{-0.1cm}
\begin{equation}
    \mathcal{L}_{GroundNet}=\mathcal{L}_{depth} + \mathcal{L}_{normal} + \eta\mathcal{L}_{seg} + \lambda\mathcal{L}_{con}
\end{equation}
where $\eta$ and $\lambda$ are hyper-parameters to controls the relative importance of the loss terms. For ground segmentation stream, we use the common softmax cross-entropy loss as $\mathcal{L}_{seg}$. Our GroundNet can be trained end-to-end.


\begin{table*}[!tb]
\setlength{\abovecaptionskip}{0.1cm}
\setlength{\belowcaptionskip}{0.1cm}
\caption{Ground plane normal evaluation results on KITTI and ApolloScape. (5\textdegree, 0.05) means the evaluation dataset is augmented by random adding roll and pitch within 5 degree and 0.05 image units, respectively. LS stands for least square module, and DSAC stands for differentiable RANSAC module. *: Note that the result of \cite{dragon2014ground} is obtained from the original published paper without any dataset augmentation, it only reports evaluation results on KITTI.} \label{eval_normal} 
\centering
\begin{tabular}{c|c|c|c|c|c|c|c}
\hline
\multicolumn{1}{c|}{\multirow{3}{*}{}} & \multicolumn{7}{c}{\textbf{error / deg}} \\ \cline{2-8} 
\multicolumn{1}{c|}{} & \multicolumn{4}{c|}{\textbf{KITTI}} & \multicolumn{3}{c}{\textbf{ApolloScape}} \\ \cline{2-8}
\multicolumn{1}{c|}{}  & 0\textdegree, 0  & 5\textdegree, 0.05 & 10\textdegree, 0.1 & 15\textdegree, 0.15 & 5\textdegree, 0.05 & 10\textdegree, 0.1 & 15\textdegree, 0.15 \\ \hline
    Marr SkipNet \cite{Bansal2016} & - & 4.95 & 7.23 & 8.89 & 5.16 & 7.33 & 8.10 \\
    GeoNet \cite{qi2018geonet} & - & 2.98 & 4.45 & 6.57 & 2.83 & 4.37 & 5.72 \\ \hline\hline
    HMM $\cite{dragon2014ground}^*$ & 4.10 & \multicolumn{3}{c|}{-} & \multicolumn{3}{c}{-}\\ \hline\hline
    \textbf{GroundNet (LS)} & 0.71 & 2.74 & 4.06 & 5.93 & 2.68 & 4.02 & 5.23 \\
    \textbf{GroundNet (DSAC)} & \textbf{0.70} & \textbf{2.65} & \textbf{3.84} & \textbf{5.41} & \textbf{2.49} & \textbf{3.72} & \textbf{4.87} \\ \hline
\end{tabular}
\vspace{-0.3cm}
\end{table*}

\begin{figure*}[!tb]
  \centering
  \setlength{\abovecaptionskip}{0cm}
  \setlength{\belowcaptionskip}{0cm}
  \includegraphics[width=1\textwidth]{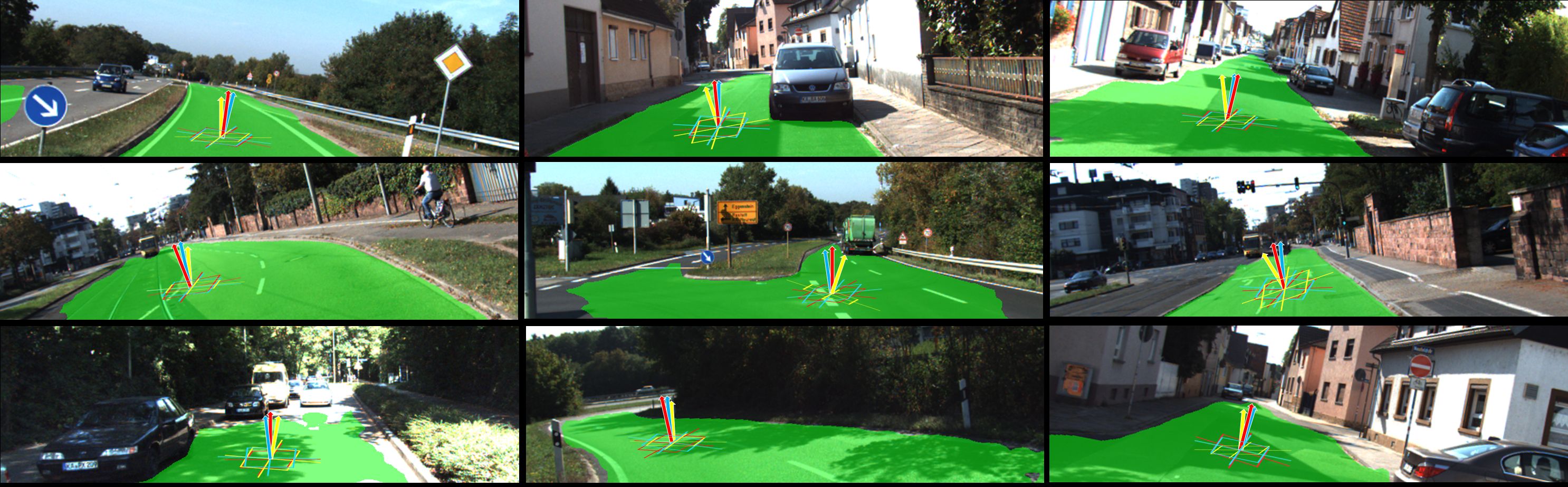}\\
  \caption{Qualitative results for the ground plane normal estimation. \textcolor{red}{Red}: groundtruth. \textcolor{blue}{Blue}: GroundNet. \textcolor{yellow}{Yellow}: Marr SkipNet. We show that our proposed GroundNet consistently outperforms all other baselines.} 
  \label{fig:normal_vis}
\end{figure*}

\section{Experiments}
\vspace{-0.1cm}
In this section, we evaluate the performance of ground plane estimation network by conducting extensive evaluations (see Sec. \ref{Normal_Evaluation} and \ref{Horizon_Evaluation}) on two augmented public benchmark datasets, KITTI and ApolloScape (see Sec. \ref{Datasets} and \ref{Augmentation}). A thorough derivation of our evaluation metrics is provided in Sec. \ref{Metrics}.  Furthermore, we perform an in-depth ablation study (see Sec. \ref{Ablation_sec}) to evaluate each component of our method. Additional details about our network architecture and training procedure are reported in Sec. \ref{Implementation}.

\subsection{Datasets} \label{Datasets}

\noindent \textbf{KITTI} is a famous and popular outdoor autonomous driving dataset \cite{uhrig2017sparsity}, in which disparity depth and road semantic labels are provided for a subset of the dataset. We adopt the split scheme proposed by Eigen et al \cite{eigen2014depth}. The ground truth of the ground plane normal is calculated from the given extrinsic matrix.

\vspace{0.2cm}
\noindent \textbf{ApolloScape} is a big autonomous driving dataset for scene parsing \cite{apolloscape_arXiv_2018}, instance segmentation and self localization. It contains a great number of image frames with complete depth information and scene labels. We use 40,963 images for training and 8330 images for validation.

\subsection{Data Augmentation} \label{Augmentation}
For both KITTI and ApolloScape, the cameras are fixed to the car, which means the ground plane orientation does not change significantly over time. As a result, the ground plane normals lack variety in the dataset. Thus, we introduce random rolls and pitches to the dataset by performing rotation and vertical translation to the images, adding variety in training and evaluation process. 

Specifically, for roll, we randomly rotate the images around their principal points within a certain limit. For pitch, we randomly translate the images up or down within a certain limit. Afterward, the images are cropped according to the principle point to get rid of black margins. This method allows us to increase the plane normal variety without introducing significant distortion or inaccuracy to the datasets.

Moreover, we set three different rotation limits (5\textdegree, 10\textdegree\ and 15\textdegree) in order to show the influence of this augmentation. Similarly, the limit of vertical translation is set as 0.05, 0.1 and 0.15 image units.

Note that the training and testing data are both augmented. Therefore, although the data is collected from a car, the testing data will possess a wide variety of orientation, significantly increasing the evaluation difficulty .

\subsection{Evaluation Metrics} \label{Metrics} 
\vspace{-0.1cm}
We compare our method with not only normal estimation methods, but also horizon line estimation methods, because horizon estimation models usually focus on outdoor scenarios. In this section, we provide two evaluation metrics with these two kinds of methods. Particularly, we give proof that under a certain assumption, the ground plane normal is equivalent to the horizon line of an image.

\vspace{0.2cm}
\noindent\textbf{Horizon Line Estimation.} To compare with horizon line estimation methods, we change our parameterization from normal vector $\mathbf{n}=[n_x,n_y,n_z]$ into $(\theta, \rho)$ and report the error of these two parameters, as mention in Sec. \ref{parameterization}. 

As shown in Figure \ref{math}, the ground plane can either be represented by its normal $\mathbf{n}$ or be represented by its roll $\theta$ and pitch $\psi$ to the horizontal plane, given the camera center and principle point (or given the intrinsic matrix $K_c$), we can transform our estimated normal into horizon line without losing any information.

From Eq. \ref{eq:pinhole} we know, a point in camera coordinate, $Q_i$, is related to a point in image coordinate $P_i$ as follows:
\vspace{-0.1cm}
\begin{equation} \label{eq:1}
    P_i=[\lambda u_i, \lambda v_i, \lambda]^T=K_cQ_i
    \vspace{-0.05cm}
\end{equation}
where $K_c$ is the intrinsic matrix. Therefore, $K_c^{-1}P_i=Q_i$. Let $P$ and $Q$ be the sets of points on the horizon line in camera and image coordinate, respectively. Then, as the normal of a plane is perpendicular to all in-plane vectors. we have $\mathbf{n} \perp Q$, or $Q^T\mathbf{n}=0$. Therefore, from equation \ref{eq:1}:
\vspace{-0.1cm}
\begin{equation} \label{eq:3}
   P^TK_c^{-T}\mathbf{n}=0
   \vspace{-0.05cm}
\end{equation}
Let $K_c^{-T}\mathbf{n}=A=[a,b,c]^T$, the horizon line in image coordinate can be represented as:
\vspace{-0.1cm}
\begin{align}
   & P^TA=0 \\
   & ax+by+cz=0
   \vspace{-0.2cm}
\end{align}
Let $z=1,  a/c=a_c, b/c=b_c$, and 
assume that the positive $x$-direction is to the right, the positive $y$-direction is down, principle point is $(W_p, H_p)$. Then we can calculate the horizon line $(\theta, \rho)$ parameters as:
\vspace{-0.2cm}
\begin{align}
    & \theta=\arctan-\frac{a_c}{b_c} \\
    & \rho=-\frac{\left|a_cW_p+b_cH_p+1\right|}{\sqrt{a_c^2+b_c^2}}
\end{align}

\vspace{-1mm}
By these derivations, the horizon line can be derived from the ground plane normal only when the intrinsic $K_c$ is available. In many use cases, such as autonomous cars or mobile robots, it is reasonable to assume that one has access to the intrinsic matrix of the camera. In both KITTI and ApolloScape datasets the camera intrinsic matrices are provided. Therefore, the estimated ground plane normal $\mathbf{n}=[n_x,n_y,n_z]^T$ can be re-parameterized as $(\theta, \rho)$ representing the horizon line in image coordinate.

\vspace{0.2cm}
\noindent\textbf{Ground Plane Normal Estimation.} To compare with other methods for ground plane normal estimation, we evaluate the angular error between the estimated ground plane normal $\mathbf{n}$ with the ground truth ground plane normal $\mathbf{n}_{gt}$ in terms of degree, as used in \cite{dragon2014ground}.

\begin{figure*}[!ht]
  \centering
  \setlength{\abovecaptionskip}{0cm}
  \setlength{\belowcaptionskip}{0cm}
  \includegraphics[width=1\textwidth]{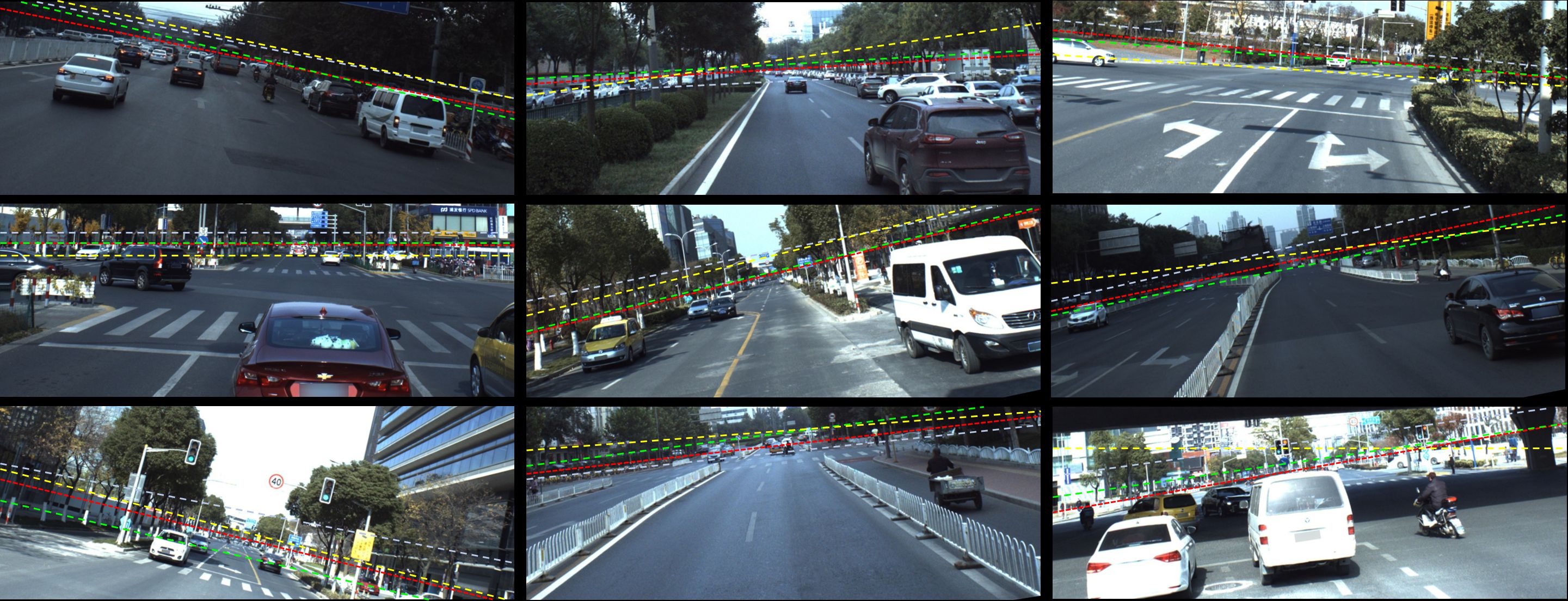}\\
  \caption{Qualitative results for the horizon line estimation. \textcolor{red}{Red}: groundtruth. \textcolor{green}{Green}: GroundNet (DSAC). \textcolor{cyan}{Cyan}: Perceptual. \textcolor{yellow}{Yellow}: DeepHorizon. \textcolor{gray}{Gray}: Zhai et al. We show that our proposed GroundNet consistently outperforms all other baselines.} \vspace{-0.3cm}
  \label{fig:horizon_vis}
\end{figure*}

\vspace{-0.1cm}
\subsection{Implementation Details} \label{Implementation}
\vspace{-0.1cm}
We implement GroundNet using the publicly available Tensorflow framework. The front-end convolutional encoder of GroundNet uses VGG-16 \cite{simonyan2014very} as backbone, plus dilated convolution and global pooling as stated in \cite{qi2018geonet}. Using deep features from the front-end encoder, we perform separate deconvolutional operations to generate three sets of task-specific feature maps. Then the separate convolutional operations are performed to produce the prediction maps. The resolution of three maps are made to be the same as input data for the followed 2D to 3D geometric operation. The surface normal map has 3 channels, including 3-direction information, and the other two maps each has 1 channel. 

We initialize the front-end encoder with network pre-trained on ImageNet. Note that as the ground segmentation task is simple, for now, we pre-train it with direct ground label supervision separately and fix its weights afterward. The weight of consistency loss $\lambda$ is set to 0.05. Adam optimizer \cite{kingma2014adam} is used to optimize the network on both datasets, $(\beta_1, \beta_2)$ is set to $(0.9, 0.999)$. For KITTI, the initial learning rate is set to 1e-4. For ApolloScape dataset, we initialize with model pre-trained on KITTI and fine-tune with learning rate set to 5e-5. We train GroundNet on one NVIDIA GTX1080Ti GPU, occupying 4GB of memory with a batch size of 4. The running time is 0.92s per image.

In practice, we set the valid depth threshold to 30m. This means only pixels with estimated depth value under 30m are used to fit the plane. This approximation prevents the far away pixels from introducing noise and biases into the estimation.

\vspace{-0.1cm}
\subsection{Ground Plane Normal Evaluation} \label{Normal_Evaluation}
\vspace{-0.1cm}
The quantitative and qualitative results for the ground plane normal estimation are shown in Table \ref{eval_normal} and Figure \ref{fig:normal_vis}. We compare our methods with two state-of-the-art surface normal estimation methods: Marr-SkipNet \cite{Bansal2016} and GeoNet \cite{qi2018geonet}; and a direct ground plane normal estimation method: HMM \cite{dragon2014ground}. 

We re-train GeoNet and Marr SkipNet on KITTI and ApolloScape with the same augmentation strategy. For GeoNet, $\{ \alpha, \beta, \gamma, \lambda\, \eta\}$ are set to $\{0.95, 9, 0.05, 0.01, 0.5\}$, with initial learning rate set to $1e^{-4}$. For Marr-SkipNet, we train the caffe model with initial learning rate set to $5e^{-4}$. As these two methods provide dense surface normal maps instead of ground plane normal vector, the final result is calculated by taking the average normal value of all pixels on the estimated ground region.

From Table \ref{eval_normal}, it is clear that GroundNet significantly outperforms all three competitors under four dataset augmentation strategies (including no augmentation). HMM is a non-learning geometry based method, which is tailored to predict ground plane normal; while other two learning-based methods are designed to predict pixel-wise surface normal map without geometry constraints. We claim that GroundNet gets better results with the integration of learning and geometry consistency.

\begin{table}[!tb]
\setlength{\abovecaptionskip}{0.1cm}
\setlength{\belowcaptionskip}{0.1cm}
\caption{Horizon line evaluation results on KITTI. (5\textdegree, 0.05) means augmentation limits as mentioned in Table \ref{eval_normal}. The unit for $\theta$ and $\rho$ are degree and $10^{-2}$ image unit.} \label{KITTI_horizon}
\begin{tabular}{c|cc|cc|cc}
\hline
\multirow{2}{*}{} & \multicolumn{2}{c|}{5\textdegree, 0.05} & \multicolumn{2}{c|}{10\textdegree, 0.1} & \multicolumn{2}{c}{15\textdegree, 0.15} \\ \cline{2-7} 
                  & $\theta$ & $\rho$ & $\theta$ & $\rho$ & $\theta$ & $\rho$ \\ \hline
Zhai et al. \cite{zhai2016detecting}  & 3.36 & 3.9 & 4.37 & 4.9 & 5.99 & 6.2\\ 
DeepHorizon \cite{workman2016hlw}    & 2.26 & 3.7 & 4.12 & 4.7 & 5.93 & 5.6\\
Perceptual \cite{Hold-Geoffroy_2018_CVPR}    & 1.98 & 3.0 & 2.94 & 3.6 & 4.66 & 4.5\\ \hline
\textbf{GroundNet}  & \textbf{1.94} & \textbf{2.6} & \textbf{2.65} & \textbf{3.1} & \textbf{4.17} & \textbf{3.8}\\ \hline
\end{tabular}
\vspace{-0.2cm}
\end{table}

\begin{table}[!tb]
\setlength{\abovecaptionskip}{0.1cm}
\setlength{\belowcaptionskip}{0.1cm}
\caption{Horizon line evaluation results on ApolloScape. (5\textdegree, 0.05) means augmentation limits as mentioned in Table \ref{eval_normal}. The unit for $\theta$ and $\rho$ are degree and $10^{-2}$ image unit.} \label{Apollo_horizon} 
\begin{tabular}{c|cc|cc|cc}
\hline
\multirow{2}{*}{} & \multicolumn{2}{c|}{5\textdegree, 0.05} & \multicolumn{2}{c|}{10\textdegree, 0.1} & \multicolumn{2}{c}{15\textdegree, 0.15} \\ \cline{2-7} 
                  & $\theta$ & $\rho$ & $\theta$ & $\rho$ & $\theta$ & $\rho$ \\ \hline
Zhai et al. \cite{zhai2016detecting}  & 2.58 & 2.9 & 3.94 & 3.7 & 5.79 & 5.1\\ 
DeepHorizon \cite{workman2016hlw}    & 1.98 & 3.2 & 3.42 & 3.3 & 4.43 & 3.9\\
Perceptual \cite{Hold-Geoffroy_2018_CVPR}  & \textbf{1.77} & 2.9 & 2.62 & 3.0 & 3.68 & 3.8\\ \hline
\textbf{GroundNet}   & 1.92 & \textbf{2.6} & \textbf{2.58} & \textbf{2.8} & \textbf{3.59} & \textbf{3.5}\\ \hline
\end{tabular}
\vspace{-0.3cm}
\end{table}

\begin{table*}[!htb]
\setlength{\abovecaptionskip}{0.1cm}
\setlength{\belowcaptionskip}{0.1cm}
\caption{Ablation results on KITTI and ApolloScape with dataset augmentation. We show the final joint training results and the stand-alone surface normal and depth streams without geometry consistency loss. LS stands for least square module.} \label{Ablation} 
\centering
\begin{tabular}{l|c|c|c|c|c|c}
\hline
\multicolumn{1}{c|}{\multirow{3}{*}{}} & \multicolumn{6}{c}{\textbf{error / deg}} \\ \cline{2-7} 
\multicolumn{1}{c|}{} & \multicolumn{3}{c|}{\textbf{KITTI}} & \multicolumn{3}{c}{\textbf{ApolloScape}} \\ \cline{2-7}
\multicolumn{1}{c|}{}                  & 5\textdegree, 0.05 & 10\textdegree, 0.1 & 15\textdegree, 0.15 & 5\textdegree, 0.05 & 10\textdegree, 0.1 & 15\textdegree, 0.15 \\ \hline
    surface normal stream  & 6.73 & 7.99 & 9.35 & 5.78 & 7.47 & 8.24 \\
    depth stream (LS) & 3.01 & 4.47 & 6.52 & 2.97 & 4.39 & 5.66 \\
    depth stream (DSAC) & 2.92 & 4.29 & 6.41 & 2.74 & 4.25 & 5.38 \\
    joint learning (LS) & 2.74 & 4.06 & 5.93 & 2.68 & 4.02 & 5.23 \\
    \textbf{joint learning (DSAC)} & \textbf{2.65} & \textbf{3.84} & \textbf{5.41} & \textbf{2.49} & \textbf{3.72} & \textbf{4.87} \\ \hline
\end{tabular}
\vspace{-0.2cm}
\end{table*}

\subsection{Horizon Line Evaluation}
\vspace{-0.1cm}
\label{Horizon_Evaluation}

Following the derivation in Sec. \ref{Metrics}, we are able to convert our estimated ground plane normal to the horizon line given the intrinsic matrix. We thus compare our method with three state-of-the-art horizon line estimation methods: Zhai et al. \cite{zhai2016detecting}, DeepHorizon \cite{workman2016hlw} and Perceptual Method \cite{Hold-Geoffroy_2018_CVPR}. Since \cite{Hold-Geoffroy_2018_CVPR} does not has the open source code, we implement a DenseNet based model according to their paper.

Quantitative results on KITTI and ApolloScape are shown in Table \ref{KITTI_horizon} and \ref{Apollo_horizon} respectively. We also plot the qualitative results in Figure \ref{fig:horizon_vis}. We find that our proposed GroundNet outperforms other methods on two datasets under all kinds of dataset augmentation. While the previous methods is able to capture the horizon line based on evident visual cues, GroundNet can get a robust estimation based on depth estimation, surface normal estimation and their intrinsic geometric relationship. Thus GroundNet gets better results when roll and pitch of the ground are steep, or when irrelevant objects (\eg, cars, trees) cover important visual cues like traffic paints and pavements.

Interestingly, the performance of \cite{Hold-Geoffroy_2018_CVPR} is close or sometimes slightly better than GroundNet when the dataset augmentation is small (\ie, 5\textdegree, 0.05). However, when we increase the amplitude of the augmentation, it is clear to see that the performance of the GroundNet gets better. This shows that our proposed method is more robust to the diversity of the rolls and pitches in the input data. Also, we notice that the average performance of all methods on ApolloScape dataset is better than that KITTI. This is because the images from the ApolloScape dataset are collected on the main road and thus contains more consistent scenarios than KITTI.

\vspace{-0.1cm}
\subsection{Ablation Study}  \label{Ablation_sec} 
\vspace{-0.1cm}
In order to justify the effect of different model components, we conduct ablation experiments for GroundNet. The results are shown in Table \ref{Ablation}. We can see that the joint training model performs better than stand-alone surface normal or depth stream. Note that the results of surface normal and depth stream are not even comparable to GeoNet, as shown in Table \ref{fig:normal_vis}. This result shows the effectiveness of our geometric consistency loss. At the same time, the differentiable RANSAC module tends to improve the results, compared with the least square module. Thus, we prove that every component of the model helps to get a better ground estimation, so the proposed GroundNet is self-consistent.


\vspace{-0.2cm}
\section{Conclusion}
\vspace{-0.1cm}
In this paper, we propose GroundNet for ground plane normal estimation from a single image. GroundNet leverages the advantages of both geometry-based and learning-based methods. A geometric consistency loss is applied to two ground plane normals computed from the surface normal stream and depth stream so that GroundNet is able to predict the consistent depth and surface normal. We tailored our model for outdoor scenarios by adding an assistive ground segmentation stream to get rid of irrelevant regions containing diverse objects and background. Experimental results on KITTI and ApolloScape datasets show that the proposed method out-performs previous state-of-the-art methods for both ground plane normal estimation and horizon line estimation in outdoor scenes.

{\small
\bibliographystyle{ieee}
\bibliography{main}
}

\end{document}